\pdfoutput=1

\documentclass[11pt]{article}

\usepackage[]{emnlp2021}

\usepackage{times}
\usepackage{latexsym}
\usepackage{amssymb}
\usepackage{multicol}
\usepackage{multirow}
\usepackage{color}
\usepackage{booktabs}
\usepackage{graphicx}
\usepackage{listings}
\usepackage{xcolor}
\usepackage{array}
\usepackage{pythonhighlight}
\usepackage{threeparttable}
\usepackage{booktabs}
\usepackage{CJKutf8}

\usepackage[T1]{fontenc}

\usepackage[utf8]{inputenc}

\usepackage{microtype}

\usepackage[textsize=scriptsize]{todonotes}

\title{WhiteningBERT: An Easy Unsupervised Sentence Embedding Approach}

\author{
Junjie Huang$^{1}\thanks{\ \ Work done during internship at microsoft.}$\hspace{0.2em}, 
Duyu Tang$^{4}$, 
Wanjun Zhong$^{2}$, 
Shuai Lu$^{3}$,\\
{\bf Linjun Shou$^{5}$,
Ming Gong$^{5}$,
Daxin Jiang$^{5}$,
Nan Duan$^{4}$}\hspace{0.5em}\\
$^{1}$Beihang University \quad
$^{2}$Sun Yat-sen University \quad
$^{3}$Peking University \\
$^{4}$Microsoft Research Asia \quad
$^{5}$Microsoft STC Asia  \\
{\tt huangjunjie@buaa.edu.cn }\\
{\tt zhongwj25@mail2.sysu.edu.cn, lushuai96@pku.edu.cn }\\
{\tt \{dutang,lisho,migon,djiang,nanduan\}@microsoft.com  }
}

\begin{document}
\maketitle
\begin{abstract}
Producing the embedding of a sentence in an unsupervised way is valuable to natural language matching and retrieval problems in practice. 
In this work, we conduct a thorough examination of pretrained model based unsupervised sentence embeddings.
We study on four pretrained models and conduct massive experiments on seven datasets regarding sentence semantics. 
We have three main findings.
First, averaging all tokens is better than only using $[CLS]$ vector.
Second, combining both top and bottom layers is better than only using top layers.
Lastly, an easy whitening-based vector normalization strategy with less than 10 lines of code consistently boosts the performance. \footnote{The whole project including codes and data is publicly available at \url{https://github.com/Jun-jie-Huang/WhiteningBERT}.}
\end{abstract}

\section{Introduction}

Pre-trained language models (PLMs) \cite{Devlin2019BERTPO, Liu2019RoBERTaAR} perform well on learning sentence semantics when fine-tuned with supervised data 
\cite{Reimers2019SentenceBERT, Thakur2020AugSBERT}. 
However, in practice, especially when a large amount of supervised data is unavailable, an approach that provides sentence embeddings in an unsupervised way is of great value in scenarios like sentence matching and retrieval.
While there are attempts on unsupervised sentence embeddings \cite{Arora2017SIF, Zhang2020ISBERT}, to the best of our knowledge, there is no comprehensive study on various PLMs with regard to multiple factors. Meanwhile, we aim to provide an easy-to-use toolkit that can be  used to produce sentence embeddings upon various PLMs. 



In this paper, we investigate PLMs-based unsupervised sentence embeddings from three aspects.
First, a standard way of obtaining sentence embedding is to pick the vector of $[CLS]$ token. We explore whether using the hidden vectors of other tokens is beneficial.
Second, some works suggest producing sentence embedding from the last layer or the combination of the last two layers \cite{Reimers2019SentenceBERT, Li2020BERTflow}. 
We seek to figure out 
whether there exists a better way of layer combination.
Third, recent attempts transform sentence embeddings to a different distribution with sophisticated networks \cite{Li2020BERTflow} to address the problem of non-smooth anisotropic distribution.
Instead, we aim to explore whether a simple linear transformation is sufficient.

To answer these questions, we conduct thorough experiments upon 4 different PLMs and evaluate
on 7 datasets regarding semantic textual similarity.
We find that, 
first, to average the token representations consistently yields better sentence representations than using the representation of the $[CLS]$ token.
Second, combining the embeddings of the bottom layer and the top layer performs than using top two layers. 
Third, normalizing sentence embeddings with whitening, an easy linear matrix transformation algorithm with less than 10 lines of code, consistently brings improvements.



\section{Transformer-based PLMs}
Multi-layer Transformer architecture \cite{Vaswani2017AttentionIA} has been widely used in pre-trained language models \citep[e.g.][]{Devlin2019BERTPO, Liu2019RoBERTaAR} to encode sentences. 
Given an input sequence $S = \{s_1, s_2, \dots, s_n\}$, a transformer-based PLM produces a set of hidden representations $H^{(0)}, H^{(1)}, \dots, H^{(L)}$, where $H^{(l)} = [\mathbf{h}^{(l)}_1, \mathbf{h}^{(l)}_2, \dots, \mathbf{h}^{(l)}_n]$ 
are the per-token embeddings of $S$ in the $l$-th encoder layer and $H^{(0)}$ corresponds to the non-contextual word(piece) embeddings. 

In this paper, we use four transformer-based PLMs to derive sentence embeddings, i.e. BERT-base \cite{Devlin2019BERTPO}, RoBERTa-base \cite{Liu2019RoBERTaAR}, DistilBERT \cite{Sanh2019DistilBERTAD}, and LaBSE  \cite{Feng2020LanguageagnosticBS}. 
They vary in the model architecture and pre-training objectives. 
Specifically, BERT-base, RoBERTa-base, and LaBSE follow an architecture of twelve layers of transformers but DistilBERT only contains six layers. 
Additionally, LaBSE is pre-trained with a unique translation ranking task which forces the sentence embeddings of a parallel sentence pair to be closer, while the other three PLMs do not include such a pre-training task for sentence embeddings.




\section{WhiteningBERT}
In this section, we introduce how to derive sentence embeddings $\mathbf{s}$ from PLMs 
following the three strategies below.


\subsection{$[CLS]$ Token v.s. Average Tokens}\label{sec:avg-tokens}
Taking the last layer of token representations as an example, we compare the following two methods to obtain sentence embeddings: (1) using the vector of $[CLS]$ token which is the first token of the sentence, i.e., $\mathbf{s}=\mathbf{s}^L=\mathbf{h}_1^L$; (2) averaging the vectors of all tokens in the sentence, including the $[CLS]$ token, i.e., $\mathbf{s}=\mathbf{s}^L=\frac{1}{n} \sum_{i=1}^N{\mathbf{h}^L_i}$.


\subsection{Layer Combination} 
Most works only take the last layer to derive sentence embeddings, 
while rarely explore which layer of semantic representations can help to derive a better sentence embedding.
Here we explore how to best combine layers of embeddings to obtain sentence embeddings.
Specifically, we can first compute the vector representation of each layer following Section \ref{sec:avg-tokens}. 
Then we perform layer combinations as $\mathbf{s}=\sum_{l} {\mathbf{s}^{l}}$ to acquire sentence embedding.
For example, for the combination of L1+L12 with two layers, we obtain sentence embeddings by adding up the vector representation of layer one and layer twelve, i.e., $\mathbf{s}=\frac{1}{2}(\mathbf{s}^{1}+\mathbf{s}^{12})$. 



\subsection{Whitening}

Whitening is a linear transformation that transforms a vector of random variables with a known covariance matrix into a new vector whose covariance is an identity matrix, and has been verified effective to improve the text representations in bilingual word embedding mapping \cite{Artetxe2018GeneralizingAI} and image retrieval \cite{Jgou2012NegativeEA}.

In our work, we explore to address the problem of non-smooth anisotropic distribution \cite{Li2020BERTflow} by a simple linear transformation method called whitening. 
Specifically, 
given a set of embeddings of $N$ sentences $\mathbf{E} = \{\mathbf{s}_1, \dots, \mathbf{s}_N \} \in \mathbb{R}^{N\times d}$, where $d$ is the dimension of the embedding,
we transform $\mathbf{E}$ linearly as in Eq. \ref{eq:white} such that $\hat{\mathbf{E}} \in \mathbb{R}^{N\times d}$ is the whitened sentence embeddings,
\begin{equation}\label{eq:white}
    \hat{\mathbf{E}} = (\mathbf{E}-m)UD^{-\frac{1}{2}},
\end{equation}
where $m \in \mathbb{R}^d$ is the mean vector of $\mathbf{E}$, $D$ is a diagonal matrix with the eigenvalues of the covariance matrix $Cov(\mathbf{E})=(\mathbf{E}-m)^T (\mathbf{E}-m) \in \mathbb{R}^{d\times d}$ and $U$ is the corresponding orthogonal matrix of eigenvectors, satisfying $Cov(\mathbf{E})=UDU^T$.

\section{Experiment}
We evaluate sentence embeddings on the task of unsupervised semantic textual similarity. 
We show experimental results and report the best way to derive unsupervised sentence embedding from PLMs.

\begin{table*}[t]
  \centering
  \resizebox{1.0\textwidth}{!}{
    \begin{tabular}{lrrrrrrr|r}
    \toprule
          Models & \multicolumn{1}{c}{STSB} & \multicolumn{1}{c}{SICK} & \multicolumn{1}{c}{STS-12} & \multicolumn{1}{c}{STS-13} & \multicolumn{1}{c}{STS-14} & \multicolumn{1}{c}{STS-15} & \multicolumn{1}{c}{STS-16} & \multicolumn{1}{c}{Avg.} \\
    \midrule
    \multicolumn{1}{l}{\textit{Baselines}} & & & & & & & & \\ 
    \multicolumn{1}{l}{\qquad Avg. GloVe \cite{Reimers2019SentenceBERT}} & 58.02 & 53.76 & 55.14 & 70.66 & 59.73 & 68.25 & 63.66 & 61.32 \\
    \multicolumn{1}{l}{\qquad SIF (GloVe+WR) \cite{Arora2017SIF}} &    -    &   -   & 56.20  & 56.60  & 68.50  & 71.70  &  -  &  63.25   \\
    \multicolumn{1}{l}{\qquad IS-BERT-NLI \cite{Zhang2020ISBERT}} & 69.21 & 64.25 & 56.77 & 69.24 & 61.21 & 75.23 & 70.16 & 66.58 \\
    \multicolumn{1}{l}{\qquad BERT-flow (NLI) \cite{Li2020BERTflow} } & 58.56 & 65.44 & 59.54 & 64.69 & 64.66 & 72.92 & 71.84 & 65.38 \\
    \multicolumn{1}{l}{\qquad SBERT WK (BERT) \cite{Wang2020SBERTWK} } & 16.07    &  41.54 & 26.66 & 14.74 & 24.32 & 28.84 & 34.37 & 26.65  \\

    \midrule
    \multicolumn{1}{l}{\textit{WhiteningBERT (PLM=\rm{BERT-base})}} & & & & & & & & \\
    \multicolumn{1}{l}{\qquad \textit{token}=CLS, \textit{layer}=L12, \textit{whitening}=F} &  20.29 & 42.42 & 32.50 & 23.99 & 28.50 & 35.51 & 51.08 & 33.47 \\
    \multicolumn{1}{l}{\qquad \textit{token}=AVG, \textit{layer}=L12, \textit{whitening}=F} & 47.29 & 58.22 & 50.08 & 52.91 & 54.91 & 63.37 & 64.94 & 55.96 \\
    \multicolumn{1}{l}{\qquad \textit{token}=AVG, \textit{layer}=L1, \textit{whitening}=F} & 58.15 & 61.78 & 58.71 & 58.21 & 62.51 & 68.86 & 67.38 & 62.23 \\
    \multicolumn{1}{l}{\qquad \textit{token}=AVG, \textit{layer}=L1+L12, \textit{whitening}=F} & 59.05 & 63.75 & 57.72 & 58.38 & 61.97 & 70.28 & 69.63 & 62.97 \\
    \multicolumn{1}{l}{\qquad \textit{token}=AVG, \textit{layer}=L1+L12, \textit{whitening}=T} & 68.68 & 60.28 & 61.94 & 68.47 & 67.31 & 74.82 & 72.82 & 67.76 \\
    \midrule
    
    \multicolumn{1}{l}{\textit{WhiteningBERT (PLM=\rm{RoBERTa-base})}} & & & & & & & & \\
    \multicolumn{1}{l}{\qquad \textit{token}=CLS, \textit{layer}=L12, \textit{whitening}=F} & 38.80 & 61.89 & 45.38 & 36.25 & 47.99 & 53.94 & 59.48 & 49.10 \\
    \multicolumn{1}{l}{\qquad \textit{token}=AVG, \textit{layer}=L12, \textit{whitening}=F} & 55.43 & 62.03 & 53.80 & 46.55 & 56.61 & 64.97 & 63.61 & 57.57 \\
    \multicolumn{1}{l}{\qquad \textit{token}=AVG, \textit{layer}=L1, \textit{whitening}=F} & 51.85 & 57.87 & 56.70 & 48.03 & 57.08 & 62.83 & 57.64 & 56.00  \\
    \multicolumn{1}{l}{\qquad \textit{token}=AVG, \textit{layer}=L1+L12, \textit{whitening}=F} & 57.54 & 60.75 & 58.56 & 50.37 & 59.62 & 66.64 & 63.21 & 59.53 \\
    \multicolumn{1}{l}{\qquad \textit{token}=AVG, \textit{layer}=L1+L12, \textit{whitening}=T} & 69.43 & 59.56 & 62.46 & 66.29 & 68.44 & 74.89 & 72.94 & 67.72 \\
    \midrule
    
    \multicolumn{1}{l}{\textit{WhiteningBERT (PLM=\rm{DistilBERT})}} & & & & & & & & \\
    \multicolumn{1}{l}{\qquad \textit{token}=CLS, \textit{layer}=L6, \textit{whitening}=F} & 30.96 & 47.73 & 40.91 & 31.30 & 39.49 & 40.64 & 57.96 & 41.29 \\
    \multicolumn{1}{l}{\qquad \textit{token}=AVG, \textit{layer}=L6, \textit{whitening}=F} & 57.17 & 63.53 & 56.16 & 59.83 & 60.42 & 67.81 & 69.01 & 61.99 \\
    \multicolumn{1}{l}{\qquad \textit{token}=AVG, \textit{layer}=L1, \textit{whitening}=F} & 55.35 & 61.34 & 57.57 & 53.79 & 60.55 & 67.06 & 63.60 & 59.89  \\
    \multicolumn{1}{l}{\qquad \textit{token}=AVG, \textit{layer}=L1+L6, \textit{whitening}=F} & 61.45 & 63.84 & 59.67 & 59.50 & 63.54 & 70.95 & 69.90 & 64.12 \\
    \multicolumn{1}{l}{\qquad \textit{token}=AVG, \textit{layer}=L1+L6, \textit{whitening}=T} & 70.37 & 58.31 & 62.09 & 68.78 & 68.99 & 75.06 & 74.52 & 68.30 \\
    \midrule
    
    \multicolumn{1}{l}{\textit{WhiteningBERT (PLM=\rm{LaBSE})}} & & & & & & & & \\
    \multicolumn{1}{l}{\qquad \textit{token}=CLS, \textit{layer}=L12, \textit{whitening}=F} & 67.18 & 69.43 & 66.99 & 61.26 & 68.36 & 77.13 & 73.10 & 69.06 \\
    \multicolumn{1}{l}{\qquad \textit{token}=AVG, \textit{layer}=L12, \textit{whitening}=F} & 71.02 & 68.36 & 67.81 & 63.94 & 70.56 & 77.93 & 75.07 & 70.67 \\
    \multicolumn{1}{l}{\qquad \textit{token}=AVG, \textit{layer}=L1, \textit{whitening}=F} & 53.70 & 55.25 & 54.81 & 44.62 & 56.97 & 60.30 & 54.57 & 54.32  \\
    \multicolumn{1}{l}{\qquad \textit{token}=AVG, \textit{layer}=L1+L12, \textit{whitening}=F} & 72.56 & 68.36 & 68.30 & 65.75 & 71.41 & 78.90 & 75.68 & 71.56 \\
    \multicolumn{1}{l}{\qquad \textit{token}=AVG, \textit{layer}=L1+L12, \textit{whitening}=T} & 73.32 & 63.27 & 68.45 & 71.11 & 71.66 & 79.30 & 74.87 & 71.71 \\
    \bottomrule
    \end{tabular}%
}
  \caption{Spearman’s rank correlation coefficient ($\rho \times 100$) between similarity scores assigned by sentence embeddings and humans. \textit{token}=AVG or \textit{token}=CLS denote using the average vectors of all tokens or only the $[CLS]$ token. L1 or L12 (L6) means using the hidden vectors of layer one or the last layer. Since DistilBERT only contains six layers of transformers, we use L6 as the last layer. T and F denote applying whitening (T) or not (F). 
  }
  \label{tab:main-results}%
\end{table*}%

\subsection{Experiment Settings}
\paragraph{Task and Datasets} The task of unsupervised semantic textual similarity (STS) aims to predict the similarity of two sentences without direct supervision. We experiment on seven STS datasets, namely the STS-Benchmark (STS-B) \cite{Cer2017STSBenchmark}, the SICK-Relatedness \cite{Marelli2014SICKR}, and the STS tasks 2012-2016 \cite{Agirre2012STS12, Agirre2013STS13, Agirre2014STS14, Agirre2015STS15, Agirre2016STS16}. These datasets consist of sentence pairs with labeled semantic similarity scores ranging from 0 to 5.

\paragraph{Evaluation Procedure} Following the procedures in previous works like SBERT \cite{Reimers2019SentenceBERT}, we first derive sentence embeddings for each sentence pair and compute the cosine similarity score of the embeddings as the predicted similarity. Then we calculate the Spearman’s rank correlation coefficient between the predicted similarity and gold standard similarity scores as the evaluation metric. We average the Spearman's coefficients among the seven datasets as the final correlation score.

\paragraph{Baseline Methods} We compare our methods with five 
representative unsupervised sentence embedding models, including average GloVe embedding \cite{Pennington2014GloveGV}, SIF \cite{Arora2017SIF} 
, IS-BERT \cite{Zhang2020ISBERT} and BERT-flow \cite{Li2020BERTflow}, SBERT-WK with BERT \cite{Wang2020SBERTWK}.

\subsection{Overall Results}

Table \ref{tab:main-results} shows the overall performance of sentence embeddings with different models and settings. 
We can observe that:

(1) Averaging the token representations of the last layer to derive sentence embeddings performs better than only using $[CLS]$ token in the last layer by a large margin, no matter which PLM we use, which indicates that single $[CLS]$ token embedding does not convey enough semantic information
as a sentence representation, despite it has been proved effective in a number of supervised classification tasks. This finding is also consistent with the results in  \citet{Reimers2019SentenceBERT}. Therefore, we suggest inducing sentence embeddings by averaging token representations.

\begin{figure*}
    \includegraphics[width=15.6cm]{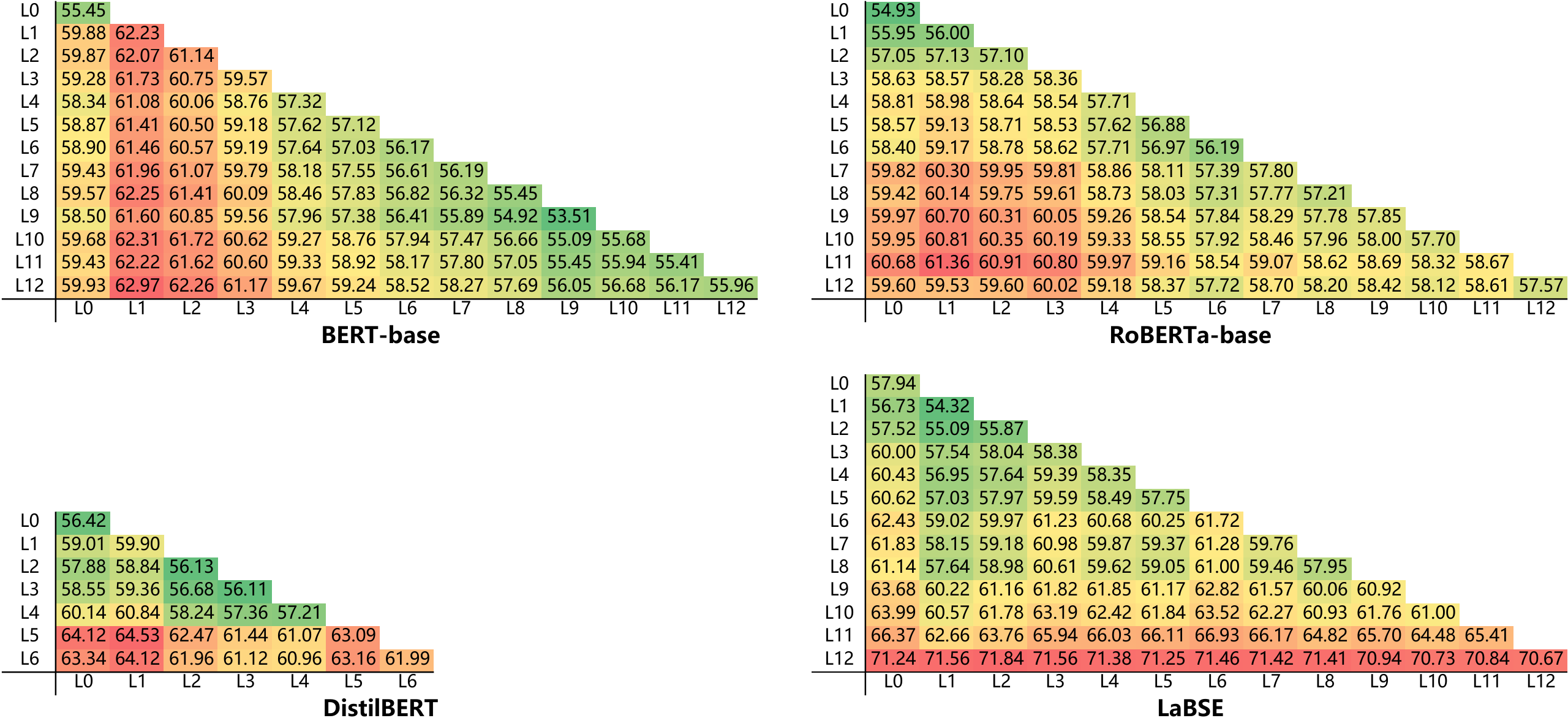}
    \caption{Performance of sentence embeddings of two layers of combinations.  X-axis and Y-axis denote the layer index. Each cell is the average correlation score of seven STS tasks of two specific layer combinations. The redder the cell is, the better performance the corresponding sentence embeddings achieve. }
    \label{fig:heatmap}
\end{figure*}

\begin{figure}[t]
    \includegraphics[width=7.8cm]{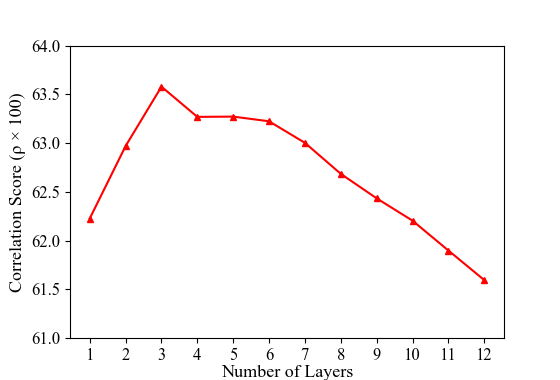}
    \caption{Maximum correlation scores of sentence embeddings from BERT-base with different numbers of combining layers. Combining three layers performs best than of other layer numbers. Especially the best combination is L1+L2+L12.}
    \label{fig:layernum}
\end{figure}

(2) Adding up the token representations in layer one and the last layer to form the sentence embeddings performs better than separately using only one layer, regardless of the selection of the PLM. Since PLMs capture a rich hierarchy of linguistic information in different layers \cite{Tenney2019BERTpipeline, Jawahar2019WhatDB}, layer combination is capable of fusing the semantic information in different layers and thus yields better performance. Therefore, we suggest summing up the last layer and layer one to perform layer combination and induce better sentence embeddings.




(3) Introducing the whitening strategy produce consistent improvement of sentence embeddings on STS tasks. 
This result indicates the effectiveness of the whitening strategy in deriving sentence embeddings. 
Among the four PLMs, LaBSE achieves the best STS performance while obtains the least performance enhancement after incorporating whitening strategy. 
We attribute it to the good intrinsic representation ability because LaBSE is pre-trained by a translation ranking task which improves the sentence embedding quality.

\subsection{Analysis of Layer Combination}
To further investigate the effects of layer combination, we add up the token representations of different layers to induce sentence embeddings. 

First, we explore whether adding up layer one and the last layer is consistently better than other combinations of two layers. 
Figure \ref{fig:heatmap} shows the performance of all two-layer combinations.
We find that adding up the last layer and layer one do not necessarily achieve the best performance among all PLMs, but could be a satisfying choice for simplicity.

Second, we explore the effects of the number of layers to induce sentence embeddings. 
We evaluate on BERT-base and figure \ref{fig:layernum} shows the maximum correlation score of each group of layer combinations. 
By increasing the number of layers, the maximum correlation score increases first but then drops. The best performance appears when the number of layers is three (L1+L2+L12). This indicates that combining three layers is sufficient to yield good sentence representations and we do not need to incorporating more layers which is not only complex but also poorly performed.

\section{Related works}
Unsupervised sentence embeddings are mainly composed with pre-trained (contextual) word embeddings \cite{Pennington2014GloveGV, Devlin2019BERTPO}. Recent attempts can be divided into two categories, according to whether the pre-trained embeddings are further trained or not.
For the former, some works leverage unlabelled natural language inference datasets to train a sentence encoder without direct supervision \cite{Li2020BERTflow, Zhang2020ISBERT, Mu2018NATSV}. 
For the latter, some works propose weighted average word embeddings based on word features \cite{Arora2017SIF, Ethayarajh2018USIF, Yang2019GEM, Wang2020SBERTWK}. However, these approaches need further training or additional features, which limits the direct applications of sentence embeddings in real-world scenarios. Finally, we note that concurrent to this work, \newcite{Su2021WhiteningSR} also explored whitening sentence embedding, released to arXiv one week before our paper.

\section{Conclusion}
In this paper, we explore different ways and find a simple and effective way to produce sentence embedding upon various PLMs. 
Through exhaustive experiments, we make three empirical conclusions here.
First, averaging all token representations consistently induces better sentence representations than using the $[CLS]$ token embedding. Second, combining the embeddings of the bottom layer and the top layer outperforms that using the top two layers. Third, normalizing sentence embeddings with a whitening algorithm consistently boosts the performance.



\bibliography{anthology,custom}
\bibliographystyle{acl_natbib}

\appendix

\section{Appendix}
\label{sec:appendix}

\begin{table*}[th]
  \centering
    \resizebox{1.0\textwidth}{!}{
    \begin{tabular}{l|c|c|c|c|c|c|c|l}
    \toprule
    PLM & STSB  & SICK  & STS-12 & STS-13 & STS-14 & STS-15 & STS-16 & Average \\
    \midrule
    BERT-base \cite{Devlin2019BERTPO} & 59.05 → 68.72 & 63.75 → 60.43 & 57.72 → 62.20 & 58.38 → 68.52 & 61.97 → 67.35 & 70.28 → 74.73 & 69.63 → 72.42 & 62.97 → 67.77 (+4.80) \\
    RoBERTa-base \cite{Liu2019RoBERTaAR} & 57.54 → 68.18 & 60.75 → 58.80 & 58.56 → 62.21 & 50.37 → 67.13 & 59.62 → 67.63 & 66.64 → 74.78 & 63.21 → 71.43 & 59.53 → 67.17 (+7.64) \\
    SpanBERT-base \cite{Joshi2019SpanBERTIP} & 59.10 → 69.82 & 60.28 → 58.48 & 58.27 → 63.16 & 54.27 → 69.00 & 61.37 → 68.71 & 67.84 → 75.37 & 66.54 → 73.24 & 61.10 → 68.25 (+7.16) \\
    DeBERTa-base \cite{He2020DeBERTaDB} & 56.55 → 67.60 & 61.66 → 59.38 & 57.55 → 62.54 & 54.78 → 67.62 & 61.43 → 66.76 & 68.84 → 74.97 & 67.51 → 71.13 & 61.19 → 67.14 (+5.95) \\
    ALBERT-base \cite{Lan2020ALBERT} & 46.18 → 61.76 & 54.99 → 58.03 & 51.02 → 58.33 & 43.94 → 62.89 & 50.79 → 59.92 & 60.83 → 68.84 & 55.35 → 65.90 & 51.87 → 62.24 (+10.37) \\
    T5-base \cite{Raffel2020ExploringTL} & 42.39 → 68.32 & 51.85 → 56.13 & 46.38 → 61.92 & 42.15 → 68.50 & 49.75 → 67.94 & 58.22 → 74.88 & 55.09 → 72.90 & 49.41 → 67.23 (+17.82) \\
    LayoutLM-base \cite{Xu2020LayoutLMPO} & 25.14 → 61.77 & 38.99 → 56.50 & 33.22 → 58.33 & 19.63 → 59.63 & 26.19 → 63.41 & 31.50 → 69.65 & 30.16 → 65.90 & 29.26 → 62.17 (+32.91) \\
    XLM-base \cite{Lample2019CrosslingualLM} & 54.47 → 69.51 & 54.65 → 55.54 & 54.52 → 62.26 & 43.15 → 66.46 & 56.50 → 69.41 & 61.10 → 75.09 & 57.30 → 73.95 & 54.53 → 67.46 (+12.93) \\
    DistilBERT \cite{Sanh2019DistilBERTAD} & 61.45 → 69.41 & 63.84 → 59.43 & 59.68 → 61.82 & 59.50 → 66.90 & 63.54 → 67.69 & 70.95 → 74.27 & 69.90 → 72.81 & 64.12 → 67.48 (+3.35) \\
    M-BERT \cite{Devlin2019BERTPO} & 57.67 → 69.09 & 58.60 → 56.85 & 58.71 → 61.13 & 53.14 → 65.74 & 61.72 → 67.18 & 68.78 → 73.64 & 67.09 → 72.53 & 60.82 → 66.60 (+5.78) \\
    MPNet \cite{Song2020MPNetMA} & 58.58 → 69.30 & 62.22 → 59.58 & 58.21 → 62.18 & 53.93 → 68.99 & 60.78 → 67.76 & 67.26 → 75.51 & 63.05 → 71.62 & 60.58 → 67.85 (+7.27) \\
    SqueezeBERT \cite{Iandola2020SqueezeBERTWC} & 54.86 → 67.80 & 60.57 → 58.43 & 56.36 → 61.43 & 53.05 → 64.57 & 60.59 → 66.96 & 67.81 → 73.57 & 64.68 → 71.24 & 59.70 → 66.29 (+6.58) \\
    LaBSE \cite{Feng2020LanguageagnosticBS} & 72.56 → 73.32 & 68.36 → 63.27 & 68.29 → 68.45 & 65.75 → 71.11 & 71.41 → 71.66 & 78.90 → 79.30 & 75.68 → 74.87 & 71.56 → 71.71 (+0.15) \\
    SPECTER \cite{Cohan2020SPECTERDR} & 62.37 → 68.90 & 57.37 → 56.42 & 62.91 → 63.62 & 52.93 → 67.43 & 62.77 → 68.82 & 67.76 → 74.47 & 66.81 → 71.04 & 61.85 → 67.24 (+5.40) \\
    MiniLM \cite{Wang2020MiniLMDS} & 50.59 → 67.91 & 58.40 → 59.79 & 55.21 → 60.32 & 44.92 → 65.00 & 54.44 → 66.35 & 64.27 → 73.79 & 59.27 → 72.38 & 55.30 → 66.51 (+11.21) \\
    \midrule
    BERT-large \cite{Devlin2019BERTPO} & 59.13 → 69.81 & 60.38 → 59.62 & 58.13 → 62.92 & 57.70 → 69.49 & 60.19 → 67.19 & 66.89 → 74.45 & 70.07 → 73.67 & 61.78 → 68.16 (+6.38) \\
    RoBERTa-large \cite{Liu2019RoBERTaAR} & 60.43 → 69.44 & 59.13 → 57.33 & 58.78 → 61.66 & 54.31 → 67.02 & 61.10 → 68.21 & 66.40 → 75.81 & 65.28 → 73.29 & 60.78 → 67.54 (+6.76) \\
    SpanBERT-large \cite{Joshi2019SpanBERTIP} & 59.51 → 70.06 & 61.10 → 58.53 & 60.85 → 63.46 & 58.36 → 71.17 & 63.24 → 69.09 & 70.43 → 75.40 & 68.24 → 73.70 & 63.10 → 68.77 (+5.67) \\
    DeBERTa-large \cite{He2020DeBERTaDB} & 57.98 → 70.28 & 62.13 → 59.11 & 58.50 → 63.48 & 55.20 → 70.10 & 62.04 → 69.10 & 70.24 → 76.76 & 68.57 → 74.56 & 62.09 → 69.06 (+6.96) \\
    ALBERT-large \cite{Lan2020ALBERT} & 50.49 → 63.45 & 57.16 → 57.98 & 55.01 → 60.29 & 49.44 → 63.15 & 53.73 → 60.81 & 65.02 → 70.16 & 60.71 → 66.37 & 55.94 → 63.17 (+7.24) \\
    T5-large \cite{Raffel2020ExploringTL}  & 35.57 → 69.16 & 40.31 → 55.75 & 37.83 → 62.33 & 29.33 → 70.70 & 39.63 → 68.41 & 45.72 → 74.82 & 47.52 → 72.01 & 39.42 → 67.60 (+28.18) \\
    LayoutLM-large \cite{Xu2020LayoutLMPO} & 45.04 → 68.16 & 49.94 → 56.32 & 49.48 → 59.50 & 32.83 → 64.28 & 42.65 → 67.60 & 47.77 → 73.14 & 49.10 → 71.81 & 45.26 → 65.83 (+20.57) \\
    XLM-large \cite{Lample2019CrosslingualLM} & 56.76 → 70.04 & 56.34 → 55.06 & 57.35 → 61.53 & 46.84 → 66.08 & 60.38 → 69.63 & 64.41 → 75.38 & 61.18 → 73.89 & 57.61 → 67.37 (+9.76) \\
    DialogRPT \cite{Gao2020DialogueRR} & 52.92 → 69.08 & 54.65 → 55.16 & 56.93 → 62.75 & 43.37 → 67.06 & 51.27 → 67.88 & 55.72 → 75.44 & 56.25 → 72.44 & 53.02 → 67.12 (+14.10) \\
    \bottomrule
    \end{tabular}
}
  \caption{ Experimental results of WhiteningBERT with different PLMs without (to the left of the arrow) or with (to the right of the arrow) whitening strategy.
  We report the Spearman’s rank correlation coefficient $(\rho \times 100)$ between similarity scores assigned by sentence embeddings and humans. 
  The embeddings are produced by averaging tokens representations (\textit{token}=AVG) and combining layer one and the last layer (\textit{layer}=L1 + L12(L24 or L6)). The average performance improves after incorporating the whitening algorithm. }
  \label{tab:add-whitening}
\end{table*}

\subsection{More Results of WhiteningBERT}
To further illustrate the effectiveness of the whitening algorithm in induce sentence embeddings for STS tasks, we experiment with more PLMs and report their performance with and without incorporating the whitening algorithm. From the results exhibited in Table \ref{tab:add-whitening}, we find that no matter which PLM we use, the average performance on 7 STS tasks improves after incorporating the whitening strategy. This result again verifies the effectiveness of whitening in producing sentence embeddings.

\subsection{Comparison with GPT-3}

GPT-3 \cite{Brown2020GPT3} is a powerful language model that is capable of sophisticated natural language understanding of tasks like classification in a zero-shot fashion. Here we report the results of whiteningBERT (PLM=BERT) on RTE dev set \cite{wang2019glue}. Specifically, we first compute the cosine similarity of the two sentence embeddings and then manually set a threshold of 0.5 to predict the label of each sentence pairs. The results are shown in Table \ref{tab:gpt3}.

\begin{table}[!t]
  \centering
   \resizebox{0.5\textwidth}{!}{
    \begin{tabular}{lrl}
    \toprule
    Model & \multicolumn{1}{l}{Accuracy} & \# Param \\
    \midrule
    GPT-3 (125M) & 47.7  & 125M \\
    GPT-3 (350M) & 49.8  & 350M \\
    GPT-3 (760M) & 48.4  & 760M \\
    GPT-3 (1.3B) & 56.0    & 1.3B \\
    GPT-3 (2.7B) & 46.6  & 2.7B \\
    GPT-3 (6.7B) & 55.2  & 6.7B \\
    GPT-3 (13B) & 62.8  & 13B \\
    GPT-3 (175B) & 63.5  & 175B \\
    \midrule
    
    whiteningBERT (PLM=BERT) & 52.7 & 110M \\
    \bottomrule
    \end{tabular}%
    }
  \caption{Experiment results on RTE. 
  The embeddings are produced by averaging tokens representations (\textit{token}=AVG), combining layer one and the last layer (\textit{layer}=L1 + L12), and incorporating whitening \textit{whitening}=T.}
  \label{tab:gpt3}%
\end{table}%

\subsection{Code for Whitening}
\begin{figure}[h]
    \includegraphics[width=7.8cm]{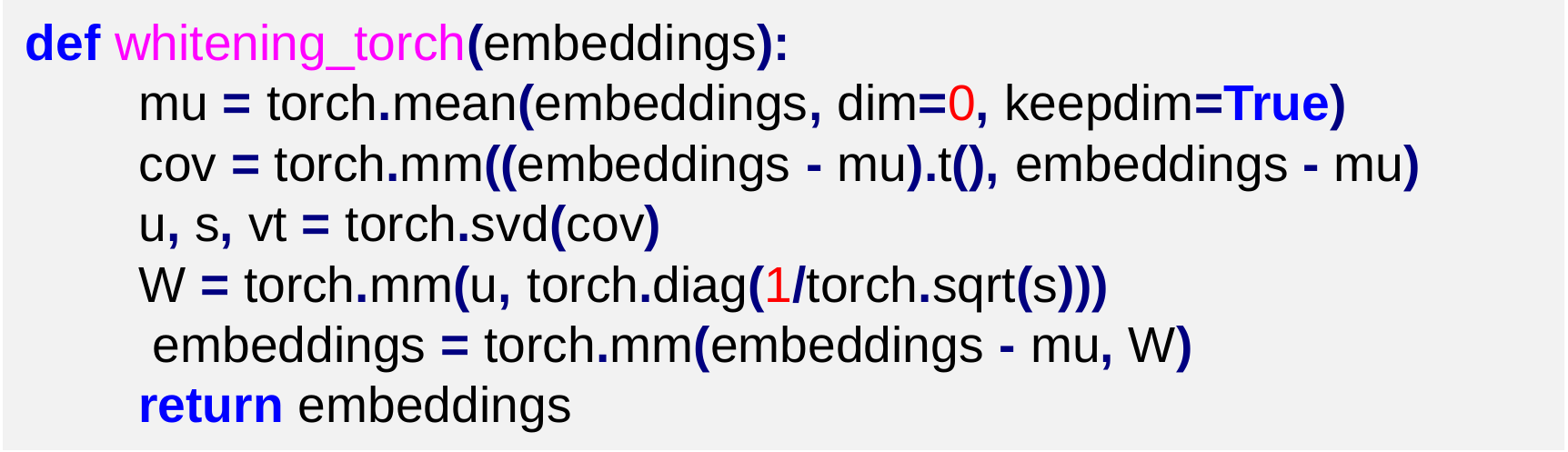}
    \caption{Pytorch code for whitening strategy.}
    \label{fig:source-code}
\end{figure}

Figure \ref{fig:source-code} displays the source code for whitening algorithm in PyTorch \cite{NEURIPS2019pytorch}.

\end{document}